\documentclass{article}

\usepackage{microtype}
\usepackage{graphicx}
\usepackage{booktabs} 
\usepackage{subcaption}
\usepackage{hyperref}
\usepackage{caption}
\usepackage{xcolor}
\usepackage[noend]{algpseudocode}
\usepackage{wrapfig}
\usepackage{adjustbox}

\usepackage{amsmath,amsfonts,bm}

\def\eqref#1{equation~\ref{#1}}

\def\1{\bm{1}}

\DeclareMathAlphabet{\mathsfit}{\encodingdefault}{\sfdefault}{m}{sl}
\SetMathAlphabet{\mathsfit}{bold}{\encodingdefault}{\sfdefault}{bx}{n}

\usepackage[accepted]{icml2019}

\newcommand{\bq}{\mathbf{q}}

\newcommand{\bx}{\mathbf{x}}
\newcommand{\by}{{y}}

\DeclareMathOperator{\dct}{DCT}

\definecolor{forestgreen}{rgb}{0.14, 0.55, 0.14}

\newcommand{\nameShort}{SimBA}
\newcommand{\nameLong}{Simple Black-box Attack}

\renewcommand{\paragraph}[1]{\vspace{0.20ex}\noindent\textbf{#1}}

\begin{document}

\twocolumn[
\icmltitle{Simple Black-box Adversarial Attacks}



\icmlsetsymbol{equal}{*}

\begin{icmlauthorlist}
\icmlauthor{Chuan Guo}{cor}
\icmlauthor{Jacob R. Gardner}{uber}
\icmlauthor{Yurong You}{cor}
\icmlauthor{Andrew Gordon Wilson}{cor}
\icmlauthor{Kilian Q. Weinberger}{cor}
\end{icmlauthorlist}

\icmlaffiliation{cor}{Department of Computer Science, Cornell University, Ithaca, New York, USA}
\icmlaffiliation{uber}{Uber AI Labs, San Francisco, California, USA}

\icmlcorrespondingauthor{Chuan Guo}{cg563@cornell.edu}

\icmlkeywords{Machine Learning, ICML}

\vskip 0.3in
]



\printAffiliationsAndNotice{}  

\begin{abstract}
We propose an intriguingly simple method for the construction of adversarial images in the black-box setting. In constrast to the white-box scenario, constructing black-box adversarial images has the additional constraint on query budget, and efficient attacks remain an open problem to date.
With only the mild assumption of continuous-valued confidence scores, our highly query-efficient algorithm utilizes the following simple iterative principle: we randomly sample a vector from a predefined orthonormal basis and either add or subtract it to the target image. Despite its simplicity, the proposed method can be used for both untargeted and targeted attacks -- resulting in previously unprecedented query efficiency in both settings. We demonstrate the efficacy and efficiency of our algorithm on several real world settings including the Google Cloud Vision API. We argue that our proposed algorithm should serve as a strong baseline for future black-box attacks, in particular because it is  extremely fast and its implementation requires less than 20 lines of PyTorch code. 
 


\end{abstract}

\section{Introduction}
As machine learning systems become prevalent in numerous application domains, the security of these systems in the presence of malicious adversaries becomes an important area of research. Many recent studies have shown that decisions output by machine learning models can be altered arbitrarily with 
imperceptible changes to the input~\citep{carlini2017towards, szegedy2013intriguing}. These attacks on machine learning models can be categorized by the capabilities of the adversary. \emph{White-box} attacks require the adversary to have complete knowledge of the target model, whereas \emph{black-box} attacks require only queries to the target model that may return complete or partial information. 

Seemingly all models for classification of natural images are susceptible to white-box attacks~\citep{athalye2018obfuscated}, which indicates that natural images tend to be close to decision boundaries learned by machine learning classifiers. Although often misunderstood as a property of neural networks~\citep{szegedy2013intriguing}, the vulnerability towards adversarial examples is likely an inevitability of classifiers in high-dimensional spaces with most data distributions~\citep{fawzi2018adversarial, shafahi2018inevitable}. 


If adversarial examples (almost) always exist, attacking a classifier turns into a search problem within a small volume around a target image. 
In the white-box scenario, this search can be guided effectively with gradient descent~\citep{szegedy2013intriguing, carlini2017towards, madry2017towards}.  
However, the black-box threat model is more applicable in many scenarios. Here, queries to the model may incur a significant cost of both time and money, and the number of black-box queries made to the model therefore serves as an important metric of efficiency for the attack algorithm. Attacks that are too costly, or are easily defeated by query limiting, pose less of a security risk than efficient attacks. To date, the average number of queries performed by the best known black-box attacks remains high despite a large amount of recent work in this area \citep{chen2017zoo, brendel2017decision, cheng2018query, guo2018low, tu2018autozoom, ilyas2018blackbox}. The most efficient and complex attacks still typically require upwards of tens or hundreds of thousands of queries. A method for query efficient black-box attacks has remained an open problem. 

Machine learning services such as Clarifai or Google Cloud Vision only allow API calls to access the model's predictions and fall therefore in the black-box category. These services do not release any internal details such as training data and model parameters; however, their predictions return continuous-valued confidence scores.  
%
%
%
In this paper we propose a simple, yet highly efficient black-box attack that exploits these confidence scores using a very simple intuition: if the distance to a decision boundary is small, we don't have to be too careful about the exact direction along which we traverse towards it. Concretely, we repeatedly pick a \textit{random} direction among a pre-specified set of orthogonal search directions, use the confidence scores to check if it is pointing towards or away from the decision boundary, and perturb the image by 
adding or subtracting the vector from the image.
Each update moves the image further away from the original image and towards the decision boundary. 

We provide some theoretical insight on the efficacy of our approach and evaluate various orthogonal search subspaces.  Similar to \citet{guo2018low}, we observe that restricting the search towards the low frequency end of the discrete cosine transform (DCT) basis is particularly query efficient. Further, we demonstrate empirically that our approach achieves a similar success rate to state-of-the-art black-box attack algorithms, however with an unprecedented low number of black-box queries. Due to its simplicity --- it can be implemented in PyTorch in under 20 lines of code\footnote{\url{https://github.com/cg563/simple-blackbox-attack}} --- we consider our method a new and perhaps surprisingly strong baseline for adversarial image attacks, and we refer to it as \textit{\nameLong{} (\nameShort{})}.

\section{Background}
The study of adversarial examples concerns with the robustness of a machine learning model to small changes in the input. The task of image classification is defined as successfully predicting what a human sees in an image. Naturally, changes to the image that are so tiny that they are imperceptible to humans should not affect the label and prediction. 
We can formalize such a robustness property as follows: 
given a model $h$ and some input-label pair $(\bx, \by)$ on which the model correctly classifies $h(\bx) = \by$, $h$ is said to be $\rho$-robust with respect to perceptibility metric $d(\cdot, \cdot)$ if 
\begin{equation*}
    h(\bx') = \by \ \ \forall \bx'\in\{\bx' \ |\  d(\bx',\bx)\leq \rho\ \}.
\end{equation*}
The metric $d$ is often approximated by the $L_0$, $L_2$ and $L_\infty$ distances to measure the degree of visual dissimilarity between the clean input $\bx$ and the perturbed input $\bx'$. Following \citep{dezfooli2016deepfool, moosavi2017universal}, for the remainder of this paper we will use $d(\bx, \bx') = \|\bx - \bx'\|_2$ as the perceptibility metric unless specified otherwise. Geometrically, the region of imperceptible changes is therefore defined to be a small hypersphere with radius $\rho$, centered around the input image $\bx$. 

Recently, many studies have shown that learned models admit directions of non-robustness even for very small values of $\rho$ \citep{dezfooli2016deepfool, carlini2017towards}. \citet{fawzi2018adversarial, shafahi2018inevitable} verified this claim theoretically by showing that adversarial examples are inherent in high-dimensional spaces. These findings motivate the problem of finding adversarial directions $\delta$ that alter the model's decision for a perturbed input $\bx' = \bx + \delta$.

\paragraph{Targeted and untargeted attacks.} The simplest success condition for the adversary is to change the original correct prediction of the model to an arbitrary class, i.e., $h(\bx') \neq \by$. This is known as an \emph{untargeted attack}. In contrast, a \emph{targeted attack} aims to construct $\bx'$ such that $h(\bx') = \by'$ for some chosen target class $\by'$. For the sake of brevity, we will focus on untargeted attacks in our discussion, but all arguments in our paper are also applicable to targeted attacks. We include experimental results for both attack types in section \ref{sec:results}. 

\paragraph{Loss minimization.} Since the model outputs discrete decisions, finding adversarial perturbations to change the model's prediction is, at first, a discrete optimization problem. However, it is often useful to define a surrogate loss $\ell_\by(\cdot)$ that measures the degree of certainty that the model $h$ classifies the input as class $\by$. The adversarial perturbation problem can therefore be formulated as the following constrained continuous optimization problem of minimizing the model's classification certainty:
\vspace{-0.5ex}
\begin{equation*}
    \min_\delta \hspace{1ex} \ell_\by(\bx + \delta) 
    \text{ subject to } \hspace{1ex} \| \delta \|_2 < \rho.
\end{equation*}
When the model $h$ outputs probabilities $p_h(\cdot \mid \bx)$ associated with each class, one commonly used adversarial loss is the probability of class $\by$: $\ell_{\by}(\bx') = p_h(\by\mid\bx')$, essentially minimizing the probability of a correct classification. For targeted attacks towards label $\by'$ a common choice is $\ell_{\by'}(\bx') = -p_h(\by'\mid\bx')$, essentially maximizing the probability of a misclassification as class $\by'$.

\paragraph{White-box threat model.} Depending on the application domain, the attacker may have various degrees of knowledge about the target model $h$. Under the \emph{white-box} threat model, the classifier $h$ is provided to the adversary. In this scenario, a powerful attack strategy is to perform gradient descent on the adversarial loss $\ell_\by(\cdot)$, or an approximation thereof. To ensure that the changees remain imperceptible, one can control the perturbation norm, $\Vert \delta \Vert_{2}$, by early stopping \citep{goodfellow2015explaining, kurakin2016adversarial} or by including the norm directly as a regularizer or constraint into the loss optimization  \citep{carlini2017towards}.


\paragraph{Black-box threat model.} Arguably, for many real-world settings the white-box assumptions may be unrealistic.  
For instance, the model $h$ may be exposed to the public as an API, allowing only queries on inputs. Such scenarios are common when attacking machine learning cloud services such as Google Cloud Vision and Clarifai. 
This \emph{black-box} threat model is much more challenging for the adversary, since gradient information may not be used to guide the finding of the adversarial direction $\delta$, and each query to the model incurs a time and monetary cost. Thus, the adversary is tasked with an additional goal of minimizing the number of black-box queries to $h$ while succeeding in constructing an imperceptible adversarial perturbation. With a slight abuse of notation this poses a modified constrained optimization problem:
%
\begin{equation*}
    \min_\delta \hspace{1ex} \ell_\by(\bx + \delta) \nonumber \\
    \text{ subject to: } \hspace{1ex} \| \delta \|_2 < \rho, 
    \text{queries} \leq B
\end{equation*}
where $B$ is some fixed budget for the number of queries allowed during the optimization. For iterative methods that query, the budget $B$ constrains the number of iterations the algorithm may take, hence requiring that the attack algorithm converges to a solution very quickly.




\section{A Simple Black-box Attack}

\begin{algorithm}[t]
\caption{\nameShort{} in Pseudocode}\label{alg:simba}
\begin{algorithmic}[1]
\Procedure{\nameShort{}($\bx,\by,Q,\epsilon$)}{}
\State $\delta= \mathbf{0}$
\State $\mathbf{p}=p_{h}(y\mid\mathbf{x})$
\While{$\mathbf{p}_y=\max_{y'}\mathbf{p}_{y'}$}
\State \textrm{Pick randomly without replacement: $\bq\in Q$ }
\For {$\alpha \in \{\epsilon,-\epsilon\}$}
\State $\mathbf{p}'=p_{h}(y\mid\bx+\delta+\alpha \bq)$
\If{$\mathbf{p}'_y<\mathbf{p}_y$}
\State $\delta=\delta+\alpha \bq$
\State $\mathbf{p}=\mathbf{p}'$
\State \textbf{break}
\EndIf{}
\EndFor{}
\EndWhile{}
\Return{$\delta$}
\EndProcedure
\end{algorithmic}
\end{algorithm}

We assume we have some image $\bx$ which a black-box neural network, $h$, classifies $h(\bx)=\by$ with predicted confidence or output probability $p_h(\by\mid\bx)$. Our goal is to find a small perturbation $\delta$ such that the prediction $h(\mathbf{x}+\delta)\neq \by$. Although gradient information is absent in the black-box setting, \textit{we argue that the presence of output probabilities can serve as a strong proxy to guide the search for adversarial images.}

\paragraph{Algorithm.} The intuition behind our method is simple (see pseudo-code in Algorithm~\ref{alg:simba}): for \emph{any} direction $\bq$ and some step size $\epsilon$, one of $\bx + \epsilon \bq$ or $\bx - \epsilon \bq$ is likely to decrease $p_{h}(y\mid\bx)$. We therefore repeatedly pick random directions $\bq$ and either add or subtract them.
To minimize the number of queries to $h(\cdot)$ we always first try adding $\epsilon\mathbf{q}$. If this decreases the probability $p_{h}(y\mid\mathbf{x})$ we take the step, otherwise we try subtracting $\epsilon \bq$. This procedure requires between 1.4 and 1.5 queries per update on average (depending on the data set and target model).
Our proposed method -- \emph{Simple Black-box Attack} (\nameShort{}) -- takes as input the target image label pair  $(\mathbf{x},y)$, a set of orthonormal candidate vectors $Q$ and a step-size $\epsilon>0$. For simplicity we pick $\mathbf{q}\in Q$ uniformly at random.
To guarantee maximum query efficiency, we ensure that no two directions cancel each other out and diminish progress, or amplify each other and increase the norm of $\delta$ disproportionately. For this reason we pick $\bq$ \textit{without replacement} and restrict all vectors in $Q$ to be \textit{orthonormal}. As we show later, this results in a guaranteed perturbation norm of  $\|\delta\|_2=\sqrt{T}\epsilon$ after $T$ updates.
The only hyper-parameters of \nameShort{}  are the set of orthogonal search vectors $Q$ and the step size $\epsilon$.

\begin{figure*}[t]
    \includegraphics[width=0.52\textwidth]{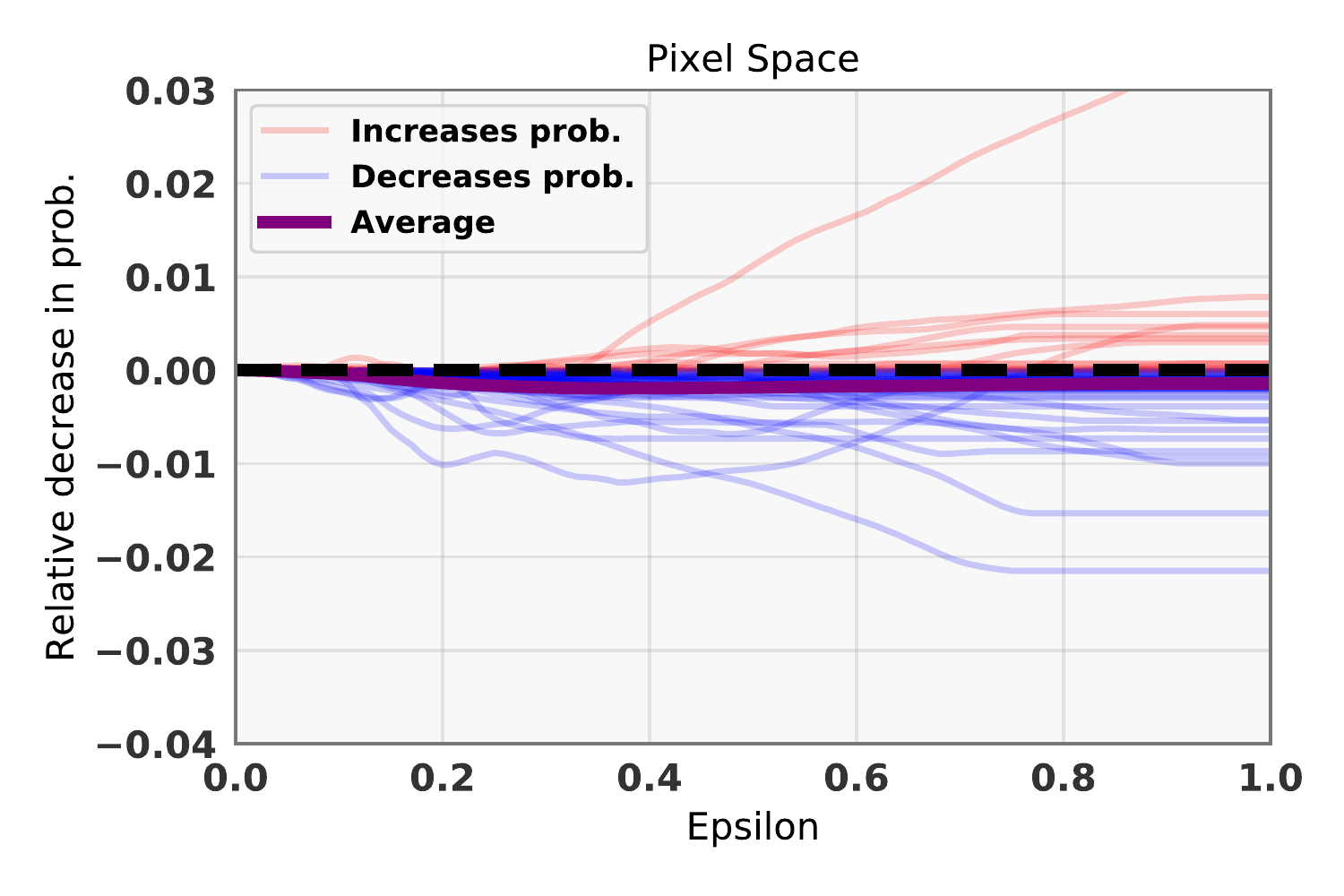}
    \includegraphics[width=0.5\textwidth]{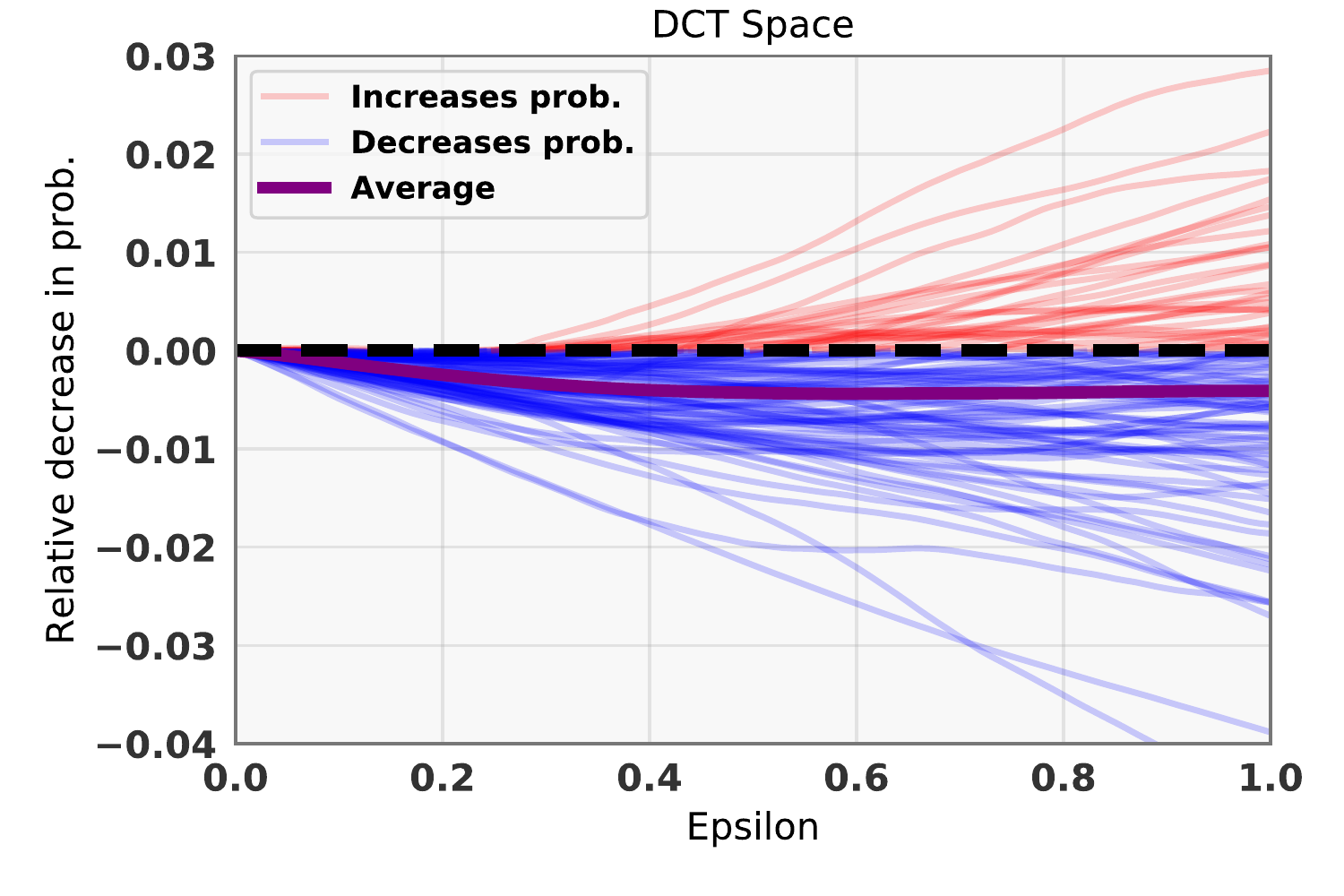}
    \vspace{-5ex}
    \caption{Plot of the change in predicted class probability when a randomly picked basis direction $\bq$ is added or subtracted (whichever decreases the loss more) with step size $\epsilon$. The left plot shows pixel space  and the right plot shows low frequency DCT space. The average change (purple line) is almost linear in $\epsilon$ with the slope being steeper when the direction is sampled in DCT space. Further, 98\% of the directions sampled in DCT space have either  $-\bq$ or $\bq$ descending, while only 73\% are descending  in pixel space.}
    \label{fig:eps_vs_prob}
    \vspace{-1ex}
\end{figure*}
%

\paragraph{Cartesian basis.}
A natural first choice for the set of orthogonal search directions $Q$ is the standard basis $Q = I$, which corresponds to performing our algorithm directly in pixel space. Essentially each iteration we are increasing or decreasing one color of a single randomly chosen pixel. Attacking in this basis corresponds to an $L_0$-attack, where the adversary aims to change as few pixels as possible.

\paragraph{Discrete cosine basis.} Recent work has discovered that random noise in low frequency space is more likely to be adversarial \citep{guo2018low}. To exploit this fact, we follow \citet{guo2018low} and propose to exploit the \emph{discrete cosine transform} (DCT). The discrete cosine transform is an orthonormal transformation that maps signals in a 2D image space $\mathbb{R}^{d \times d}$ to frequency coefficients corresponding to magnitudes of cosine wave functions. In what follows, we will refer to the set of orthonormal frequencies extracted by the DCT as $Q_{\dct}$. While the full set of directions $Q_{\dct}$ contains $d \times d$ frequencies, we keep only a fraction $r$ of the lowest frequency directions in order to make the adversarial perturbation in the low frequency space.

\paragraph{General basis.} In general, we believe that our attack can be used with any orthonormal basis, provided that the basis vectors can be sampled efficiently. This is especially challenging for high resolution datasets such as ImageNet since each orthonormal basis vector has dimensionality $d \times d$. Iterative sampling methods such as Gram-Schmidt process cannot be used due to linear memory cost in the number of sampled vectors. Thus, we choose to evaluate our attack using only the standard basis vectors and DCT basis vectors for their efficiency and natural suitability to images.

\paragraph{Learning rate $\epsilon$.}
Given any set of search directions $Q$, some directions may decrease $p_h(y\mid\bx)$ more than others. Further, it is possible for the output probability $p_{h}(y\mid\bx + \epsilon \bq)$ to be non-monotonic in $\epsilon$.
In \autoref{fig:eps_vs_prob}, we plot the relative decrease in probability as a function of $\epsilon$ for randomly sampled search directions in both pixel space and the DCT space. The probabilities correspond to prediction on an ImageNet validation sample by a ResNet-50 model. This figure highlights an illuminating result:
the probability $p_h(y\mid\mathbf{x}\pm\epsilon\bq)$  \emph{decreases monotonically} in $\epsilon$ with surprising consistency (across random images and vectors $\bq$)!
Although some directions eventually increase the true class probability, the expected change in this probability is negative with a relatively steep slope.
This means that our algorithm is not overly sensitive to the choice of $\epsilon$ and the iterates will decrease the true class probability quickly. The figure also shows that search in the DCT space tends to lead  to steeper descent directions than pixel space.
As we show in the next section, we can tightly bound the final $L_2$-norm of the perturbation given a choice of $\epsilon$ and maximum number of steps $T$, so the choice of $\epsilon$ depends \emph{primarily on budget considerations} with respect to $\|\delta\|_2$.

\paragraph{Budget considerations.}
%
By exploiting the orthonormality of the basis $Q$ we can bound the norm of $\delta$ tightly. Each iteration a basis vector is either added, subtracted, or discarded (if neither direction yields a reduction of the output probability.)
Let $\alpha_i \in \{-\epsilon, 0, \epsilon\}$ denote the sign of the search direction chosen at step $t$, so
\begin{equation*}
    \delta_{t + 1} = \delta_{t} + \alpha_t\bq_{t}.
\end{equation*}
We can recursively expand $\delta_{t + 1} = \delta_{t } + \alpha_t\bq_{t}$. In general, the
final perturbation $\delta_{T}$ after $T$ steps can be written as a sum of these individual search directions:
\begin{equation*}
    \delta_{T} = \sum_{t=1}^{T} \alpha_t \bq_{t}.
\end{equation*}
Since the directions $\bq_{t}$ are orthogonal, $\bq_{t}^{\top}\bq_{t'} = 0$ for any $t \neq t'$. We can therefore compute the $L_2$-norm of the adversarial perturbation:
\begin{align*}
    \Vert \delta_{T} \Vert_2^2 = \left\| \sum_{t=1}^{T} \alpha_t \bq_{t} \right\|_2^2
                          = \sum_{t=1}^{T} \Vert \alpha_i \bq_{t} \Vert_2^2
													 &= \sum_{t=1}^{T} \alpha_t^2 \Vert \bq_{t} \Vert_2^2 \\
                          &\leq T\epsilon^2.
\end{align*}
Here, the second equality follows from the orthogonality of $\bq_{t}$ and $\bq_{t'}$, and the last inequality is tight if all queries result in a step of either $\epsilon$ or $-\epsilon$. Thus the adversarial perturbation has $L_2$-norm at most $\sqrt{T} \epsilon$ after $T$ iterations. This result holds for any orthonormal basis (e.g. $Q_{\dct}$).

Our analysis highlights an important trade-off: for query-limited scenarios, we may reduce the number of iterations by setting $\epsilon$ higher, incurring higher perturbation $L_2$-norm. If a low norm solution is more desirable, reducing $\epsilon$ will allow quadratically more queries at the same $L_2$-norm. A more thorough theoretical analysis of this trade-off could improve query efficiency.

\section{Experimental Evaluation}
\label{sec:results}
In this section, we evaluate our attack against a comprehensive list of competitive black-box attack algorithms: the Boundary Attack \citep{brendel2017decision}, Opt attack \citep{cheng2018query}, Low Frequency Boundary Attack (LFBA) \citep{guo2018low}, AutoZOOM \citep{tu2018autozoom}, the QL attack \citep{ilyas2018blackbox}, and the Bandits-TD attack \citep{ilyas2018prior}. There are three dimensions to evaluate black-box adversarial attacks on: how often the optimization problem finds a feasible point (\emph{success rate}), how many queries were required ($B$), and the resulting perturbation norms  $(\rho)$.

\begin{figure*}
\centering
\includegraphics[width=0.49\textwidth]{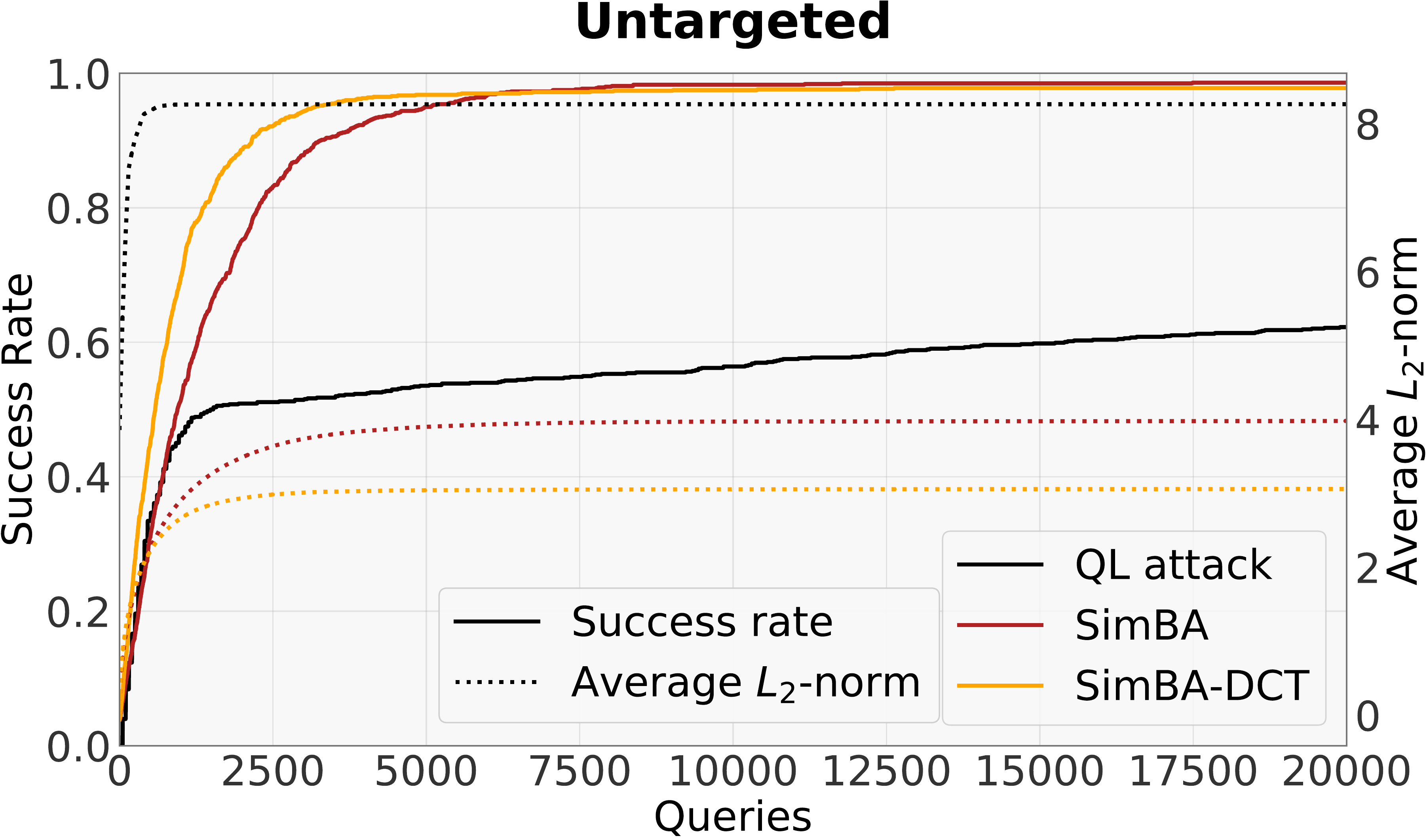}
\includegraphics[width=0.49\textwidth]{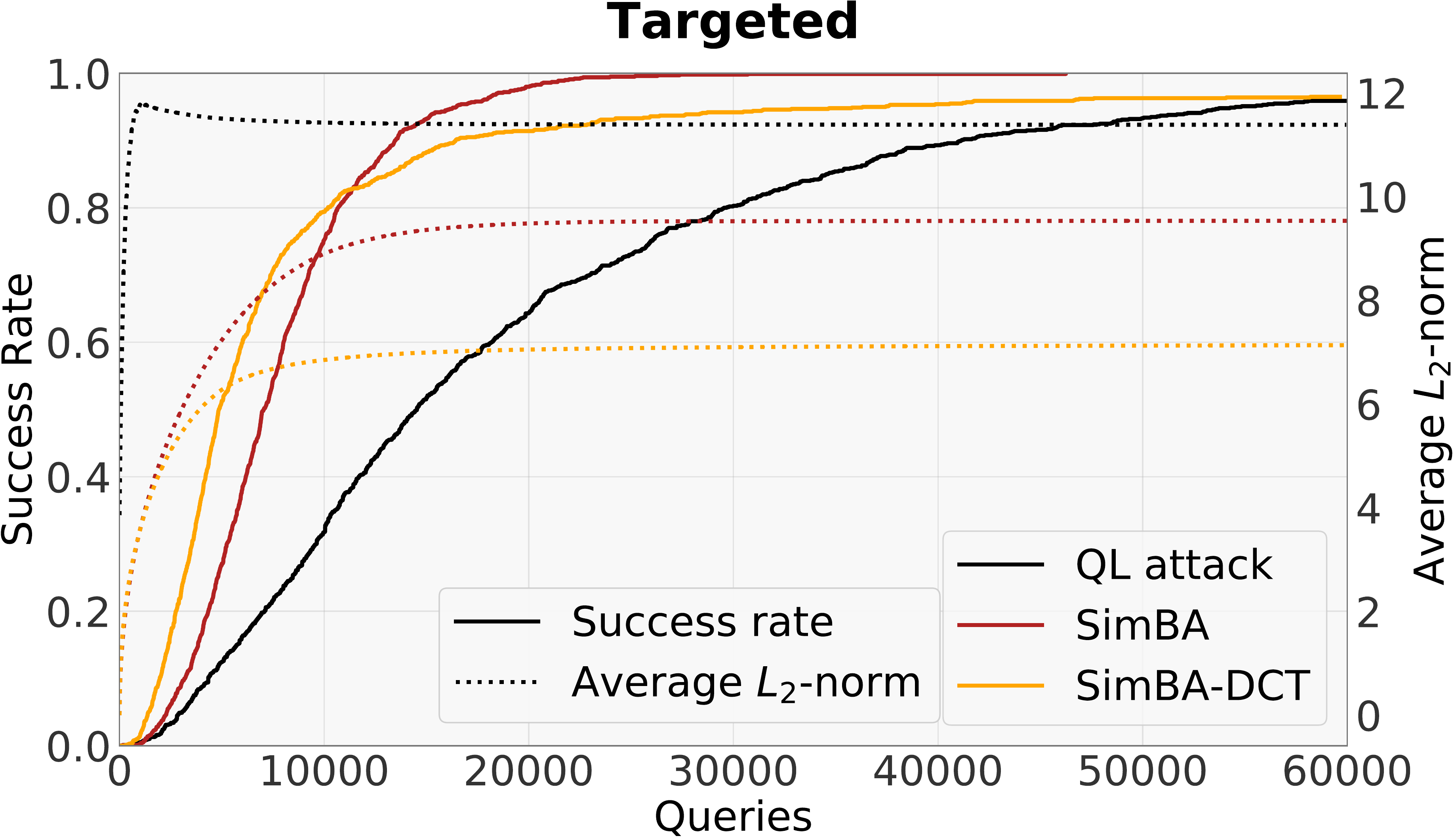}
\vspace{-1ex}
\caption{Comparison of success rate and average $L_2$-norm versus number of model queries for untargeted (left) and targeted (right) attacks. Horizontal axis shows number of model queries. The increase in success rate for SimBA and SimBA-DCT are dramatically faster than that of QL-attack in both untargeted and targeted scenarios. Both methods also achieve lower average $L_2$-norm than QL-attack. Note that although SimBA-DCT has faster initial convergence, its final success rate is lower than SimBA.
\label{fig:methods_comparison}}
\end{figure*}


\subsection{Setup}


We first evaluate our method on ImageNet. We sample a set of 1000 images from the ImageNet validation set that are initially classified correctly to avoid artificially inflating the success rate. Since the predicted probability is available for every class, we minimize the probability of the correct class as adversarial loss in untargeted attacks, and maximize the probability of the target class in targeted attacks. We sample a target class uniformly at random for all targeted attacks.

Next, we evaluate SimBA in the real-world setting of attacking the Google Cloud Vision API. Due to the extreme budget required by baselines that might cost up to \$150 per image\footnote{The Google API charges \$1.50 for 1000 image queries.}, here we only compare  to LFBA, which we found to be the most query efficient baseline.

In our experiments, we limit SimBA and SimBA-DCT to at most $T=10,000$ iterations for untargeted attacks and to $T=30,000$ for targeted attacks. For SimBA-DCT, we keep the first $1/8$th of all frequencies, and add an additional $1/32$nd of the frequencies whenever we exhaust available frequencies without succeeding. For both methods, we use a fixed step size of $\epsilon = 0.2$.

\subsection{ImageNet results}
\label{sec:imagenet}

\paragraph{Success rate comparison (\autoref{fig:methods_comparison}).}
We demonstrate the query efficiency of our method in comparison to the QL attack -- arguably the state-of-the-art black-box attack method to date -- by plotting the average success rate against the number of queries. Figure \ref{fig:methods_comparison} shows the comparison for both untargeted and targeted attacks. The dotted lines show progression of average $L_2$-norm throughout optimization. Both SimBA and SimBA-DCT achieve dramatically faster increase in success rate in both untargeted and targeted scenarios. The average $L_2$-norm for both methods are also significantly lower.

\begin{figure*}[t!]
\centering
\includegraphics[width=0.49\textwidth]{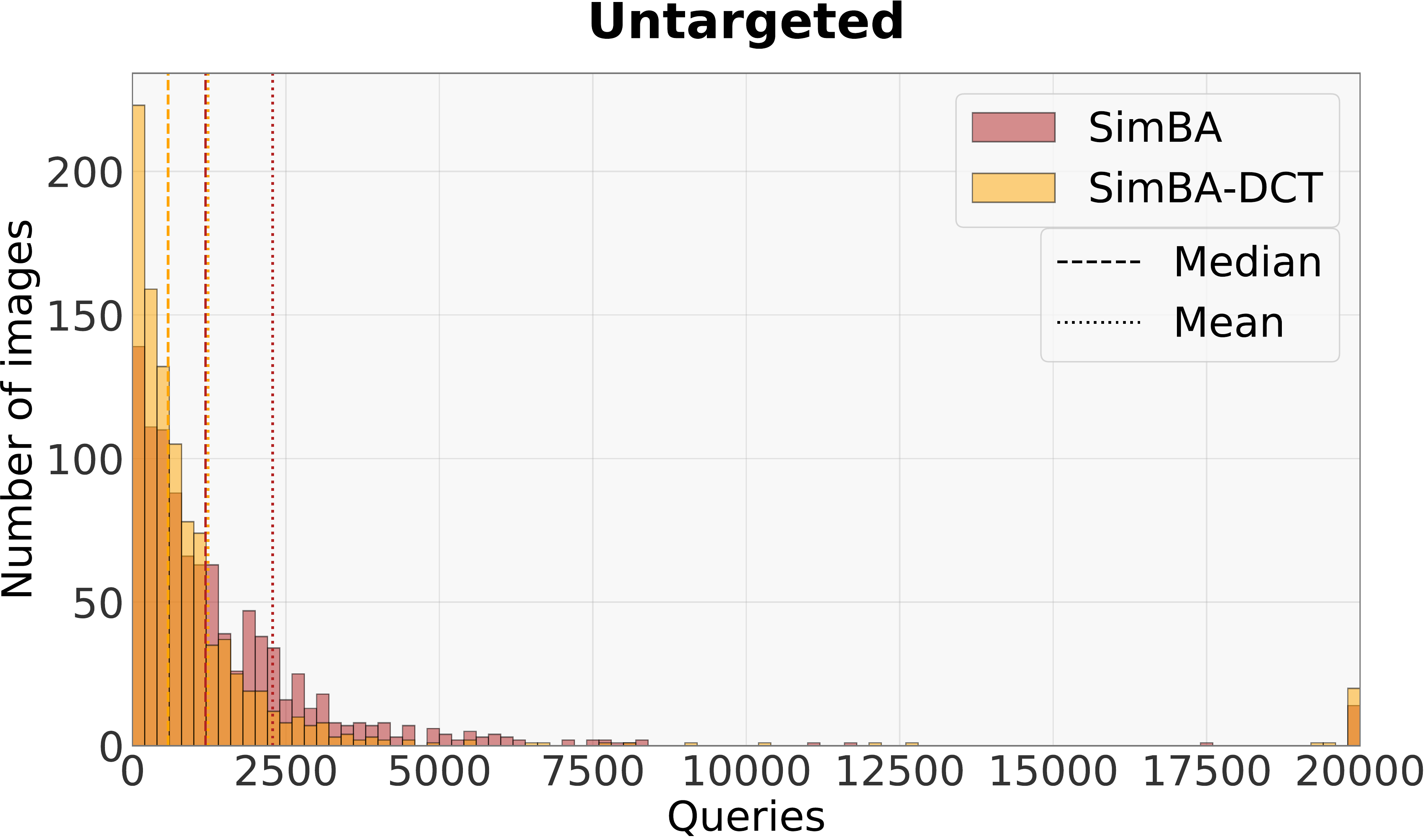}
\includegraphics[width=0.49\textwidth]{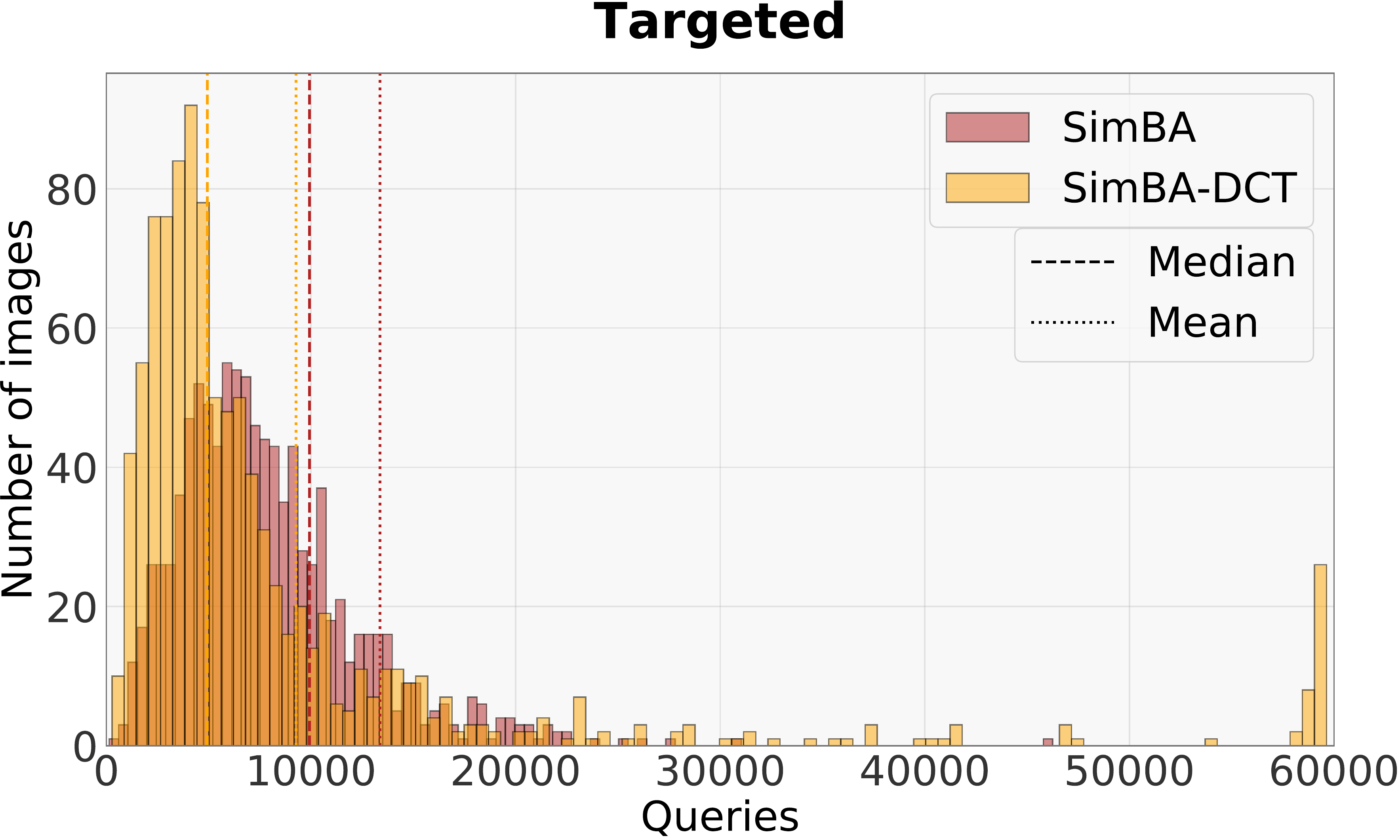}
\vspace{-1ex}
\caption{Histogram of number of queries required until a successful attack (over 1000 target images). SimBA-DCT is highly right skewed, suggesting that only a handful of images require more than a small number of queries. The \emph{median} number of queries required by SimBA-DCT for untargeted attack is only 582. However, limiting to the low frequency basis results in SimBA-DCT failing to find a successful adversarial image after $60,000$ queries, whereas SimBA can achieve $100\%$ success rate consistently.
\label{fig:queries_histogram}}
\end{figure*}

\begin{table*}[ht!]
\centering
\resizebox{0.49\textwidth}{!}{
\begin{tabular}{cccc}
\multicolumn{4}{c}{\large \hspace{4ex} \textbf{Untargeted}} \\
\hline
\textbf{Attack} & \textbf{Average queries} & \textbf{Average $L_2$} & \textbf{Success rate}  \\
\hline
\multicolumn{4}{c}{\hspace{4ex} Label-only} \\
\hline
Boundary attack & 123,407 & 5.98 & $100\%$ \\
Opt-attack & 71,100 & 6.98 & $100\%$ \\
LFBA & 30,000 & 6.34 & $100\%$ \\
\hline
\multicolumn{4}{c}{\hspace{4ex} Score-based} \\
\hline
QL-attack & 28,174 & 8.27 & $85.4\%$ \\
Bandits-TD & 5,251 & 5.00 & $80.5\%$ \\
\textbf{SimBA} & 1,665 & 3.98 & $98.6\%$ \\
\textbf{SimBA-DCT} & {\bf 1,283} & 3.06 & $97.8\%$ \\
\hline
\end{tabular}
}
\resizebox{0.49\textwidth}{!}{
\begin{tabular}{cccc}
\multicolumn{4}{c}{\large \hspace{3ex} \textbf{Targeted}} \\
\hline
\textbf{Attack} & \textbf{Average queries} & \textbf{Average $L_2$} & \textbf{Success rate} \\
\hline
\multicolumn{4}{c}{\hspace{4ex} Score-based} \\
\hline
QL-attack & 20,614 & 11.39 & $98.7\%$ \\
AutoZOOM & 13,525 & 26.74 & $100\%$ \\
\textbf{SimBA} & {\bf 7,899} & 9.53 & $100\%$ \\
\textbf{SimBA-DCT} & 8,824 & 7.04 & $96.5\%$ \\
\hline
\end{tabular}
}
\vspace{1ex}
\caption{Average query count for untargeted (left) and targeted (right) attacks on ImageNet. Methods are evaluated on three different metrics: average number of queries until success (lower is better), average perturbation $L_2$-norm (lower is better), and success rate (higher is better). Both SimBA and SimBA-DCT achieve close to $100\%$ success rate, similar to other methods in comparison, but require significantly fewer model queries while achieving \emph{lower} average $L_2$ distortion.
\label{table:method_comparison_imagenet}}
\end{table*}

\paragraph{Query distributions (\autoref{fig:queries_histogram}).} In \autoref{fig:queries_histogram} we plot the histogram of model queries made by both SimBA and SimBA-DCT over 1000 random images. Notice that the distributions are highly right skewed so the median query count is a much more representative aggregate statistic than average query count. These median counts for SimBA and SimBA-DCT are only 944 and 582, respectively. In the targeted case, SimBA-DCT can construct an adversarial perturbation within only $4,854$ median queries but failed to do so after $60,000$ queries for approximately $2.5\%$ of the images. In contrast, SimBA achieves a success rate of $100\%$ with a median query count of $7,038$.

This result shows a fundamental trade-off when selecting the orthonormal basis $Q$. Restricting to only the low frequency DCT basis vectors for SimBA-DCT results in faster average rate of descent for most images, but may fail to admit an adversarial perturbation for some images. This phenomenon has been observed by \citet{guo2018low} for optimization-based white-box attacks as well. Finding the right spectrum to operate in on a per-image basis may be key to further improving the query efficiency and success rate of black-box attack algorithms. We leave this promising direction for future work.

\paragraph{Aggregate statistics (\autoref{table:method_comparison_imagenet}).}
\autoref{table:method_comparison_imagenet} computes aggregate statistics of model queries, success rate, and perturbation $L_2$-norm across different attack algorithms. We reproduce the result for LFBA,  QL-attack and Bandits-TD using default hyperparameters, and present numbers reported by the original authors' papers for Boundary Attack\footnote{Result reproduced by \citet{cheng2018query}}, Opt-attack, and AutoZOOM. The target model is a pretrained ResNet-50 \citep{he2016residual} network, with the exception of AutoZOOM, which used an Inception v3 \citep{szegedy2016rethinking} network. Some of the attacks operate under the harder label-only setting (i.e., only the predicted label is observed), which may impact their query efficiency due to observation of partial information. Nevertheless, we include these methods in the table for completeness.

The three columns in the table show all the relevant metrics for evaluating a black-box attack. Ideally, an attack should succeed often, construct perturbation with low $L_2$ norm, and do so within very few queries. It is possible to artificially reduce the number of model queries by lowering success rate and/or increasing perturbation norm. To ensure fair comparison, we enforce our methods achieve close to $100\%$ success rate and compare the other two metrics. Note that the success rate for boundary attack and LFBA are always $100\%$ since both methods begin with very large perturbations to guarantee misclassification and gradually reduce the perturbation norm.

Both SimBA and SimBA-DCT have \emph{significantly lower} average $L_2$-norm than all baseline methods. For untargeted attack, our methods require 3-4x fewer queries (at $1,665$ and $1,232$, respectively) compared to the strongest baseline method -- Bandits-TD -- which only achieves a $80\%$ success rate. For targeted attack (right table), the evaluated methods are much more comparable, but both SimBA and SimBA-DCT still require significantly fewer queries than QL-attack and AutoZOOM.

\begin{figure}[t]
\vspace{1ex}
\centering
\includegraphics[width=\columnwidth]{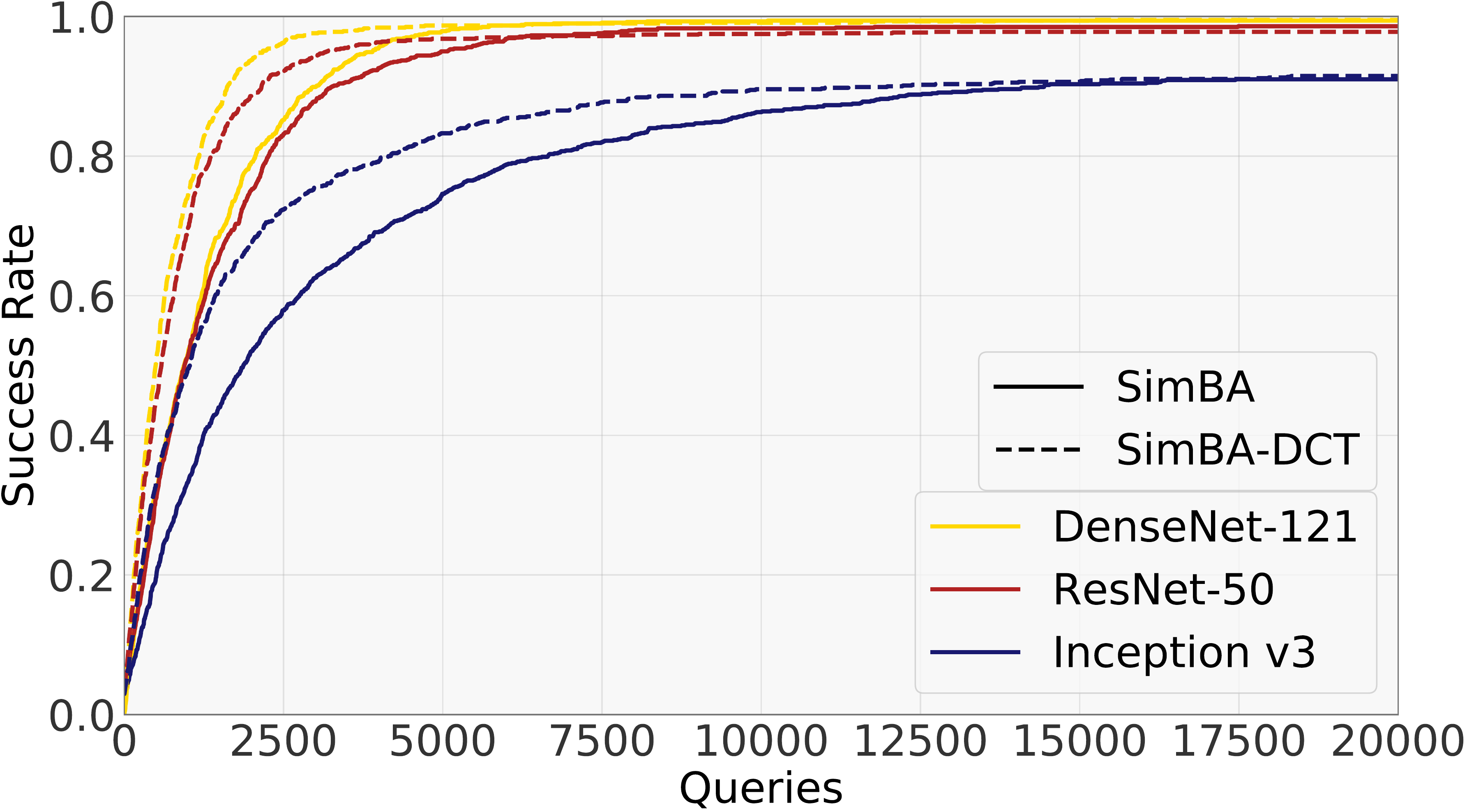}
\caption{Comparison of success rate versus number of model queries across different network architectures for untargeted SimBA (solid line) and SimBA-DCT (dashed line) attacks. Both methods can successfully construct adversarial perturbations within $20,000$ queries with high probability. DenseNet is the most vulnerable against both attacks, admitting a success rate of almost $100\%$ after only 6,000 queries for SimBA and 4000 queries for SimBA-DCT. Inception v3 is much more difficult to attack for both methods.
\label{fig:network_comparison}}
\vspace{1ex}
\end{figure}

\paragraph{Evaluating different networks (\autoref{fig:network_comparison}).}
To verify that our attack is robust against different model architectures, we evaluate SimBA and SimBA-DCT additionally against DenseNet-121 \citep{huang2016densely} and Inception v3 \citep{szegedy2016rethinking} networks. Figure \ref{fig:network_comparison} shows success rate across the number of model queries for an untargeted attack against the three different network architectures. ResNet-50 and DenseNet-121 exhibit a similar degree of vulnerability against our attacks. However, Inception v3 is noticeably more difficult to attack, requiring more than $10,000$ queries to successfully attack with some images. Nevertheless, both methods can successfully construct adversarial perturbations against all models with high probability.

\begin{figure}[ht!]
\centering\includegraphics[width=1.08\columnwidth]{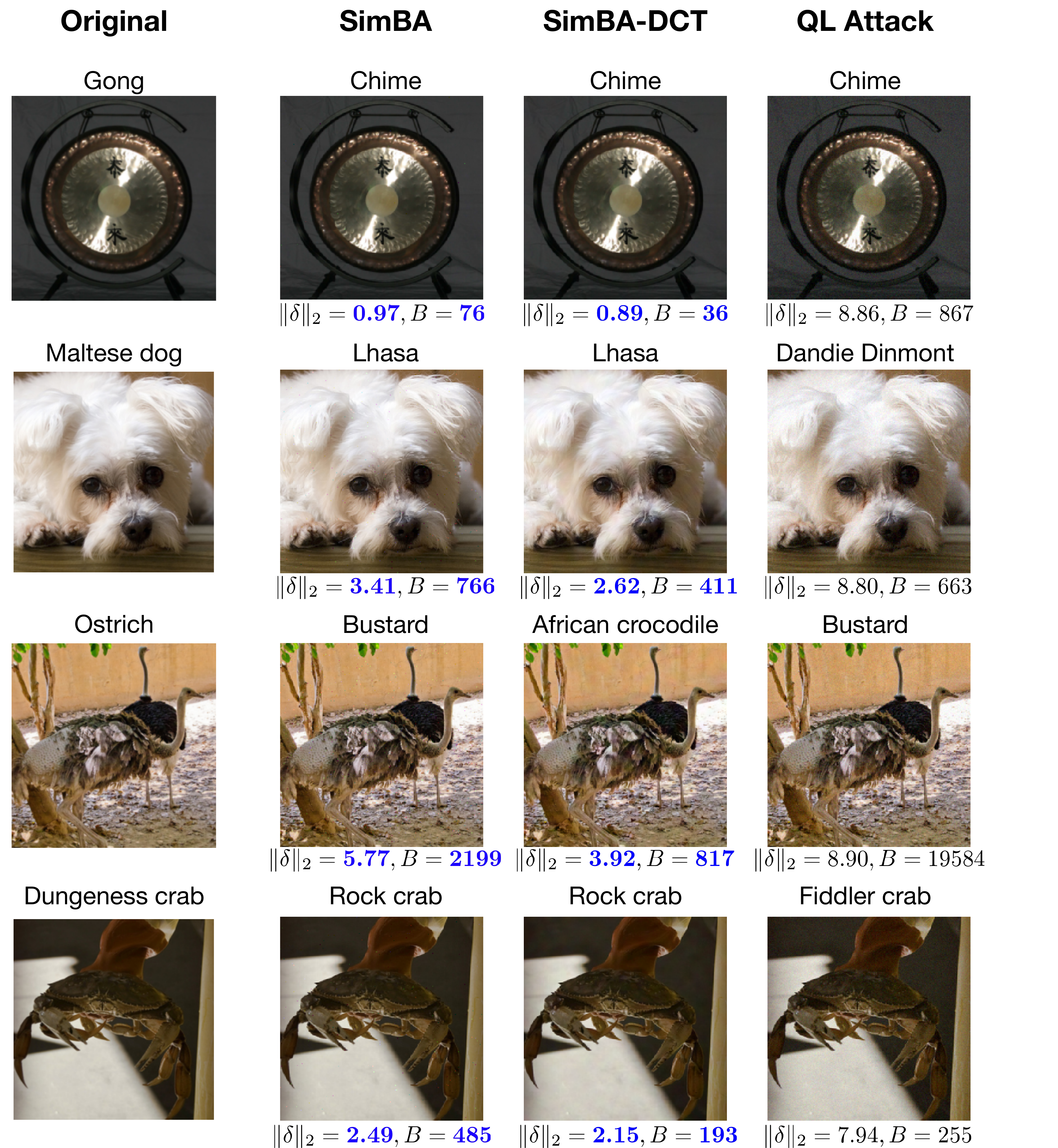}
\caption{Randomly selected images before and after adversarial perturbation by SimBA, SimBA-DCT and QL attack. The constructed perturbation is imperceptible for all three methods, but the perturbation $L_2$-norms for SimBA and SimBA-DCT are significantly lower than that of QL attack across all images. Our methods are capable of constructing an adversarial example in comparable or fewer queries than QL attack -- as few as 36 queries in some cases! Zoom in for detail. \label{fig:image_samples}}
\vspace{-2ex}
\end{figure}

\begin{figure}[t]
    \centering
    \vspace{1ex}
    \includegraphics[width=0.8\columnwidth]{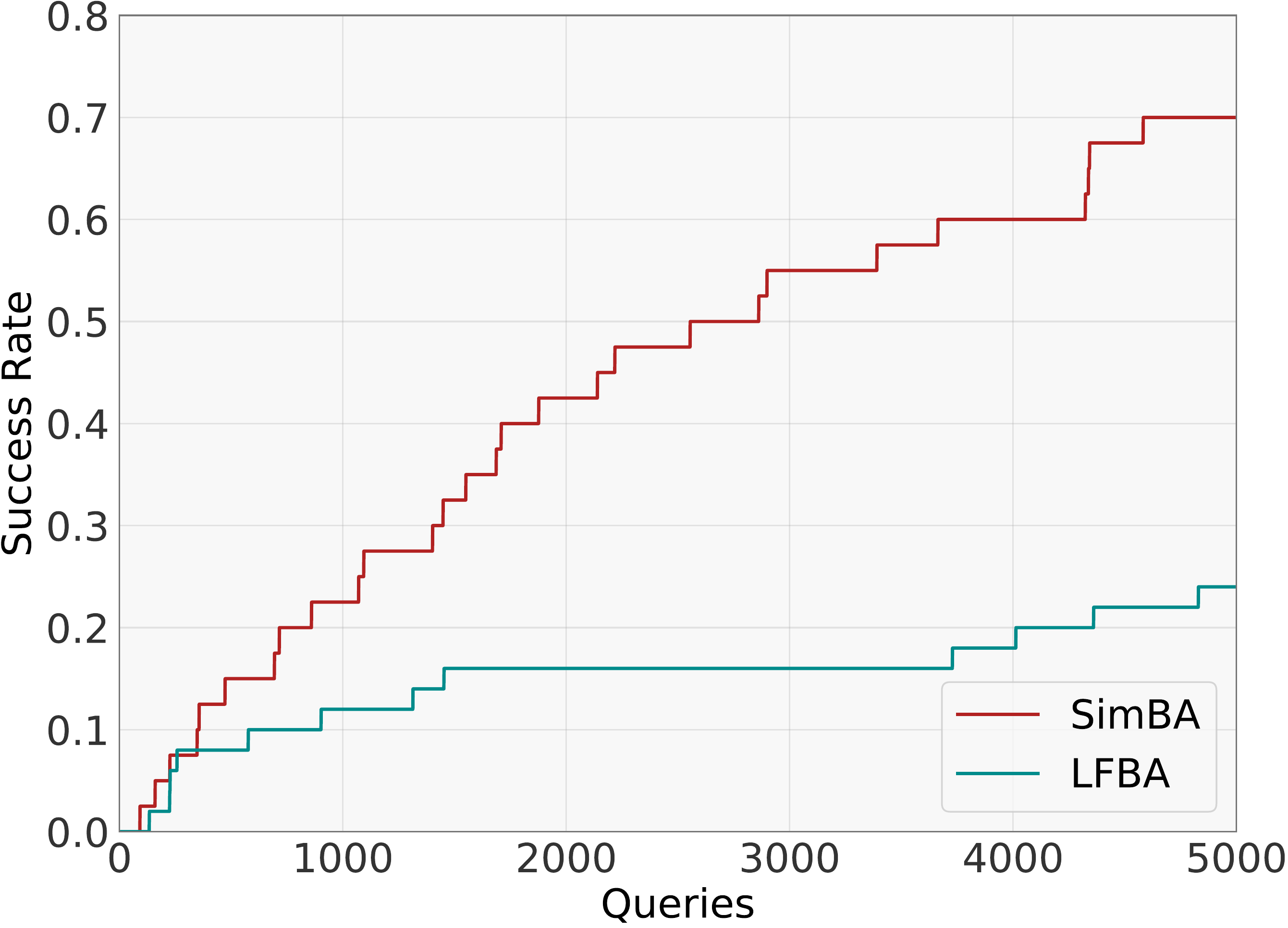}
    \vspace{-1ex}
    \caption{Plot of success rate across number of model queries for Google Cloud Vision attack. SimBA is able to achieve close to $70\%$ success rate after only 5000 queries, while the success rate for LFBA has only reached $25\%$. \label{fig:gcv_succ_plot}}
    \vspace{-3ex}
\end{figure}

\begin{figure*}[t]
    \centering
    \includegraphics[width=\textwidth]{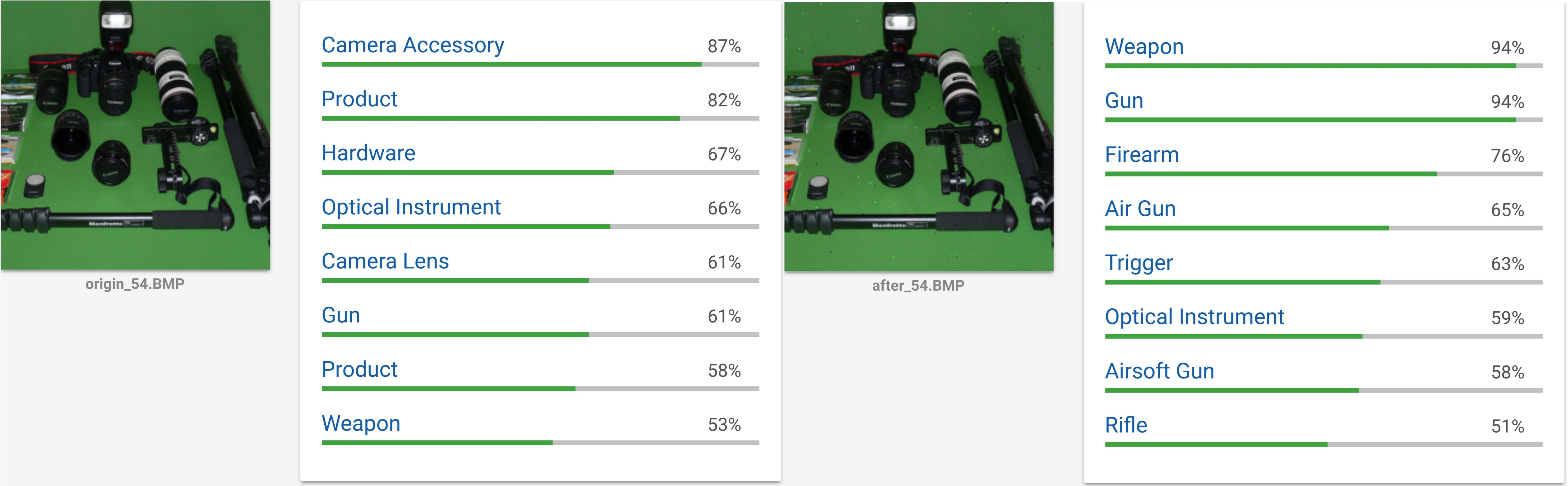}
    \caption{Screenshot of Google Cloud Vision labeling results on a sample image before and after adversarial perturbation. The original image contains a set of camera instruments. The adversarial image successfully replaced the top concepts with guns and weapons. See supplementary material for additional samples. \label{fig:gcv_screenshot}}
    \vspace{-2ex}
\end{figure*}

\paragraph{Qualitative results (\autoref{fig:image_samples}).}
For qualitative evaluation of our method, we present several randomly selected images before and after adversarial perturbation by untargeted attack. For comparison, we attack the same set of images using QL attack. Figure \ref{fig:image_samples} shows the clean and perturbed images along with the perturbation $L_2$-norm and number of queries. While all attacks are highly successful at changing the label, the norms of adversarial perturbations constructed by SimBA and SimBA-DCT are much smaller than that of QL attack. Both methods requires consistently fewer queries than QL attack for almost all images. In fact, SimBA-DCT was able to find an adversarial image in as few as 36 model queries! Notice that the perturbation produced by SimBA contains sparse but sharp differences, constituting a low $L_0$-norm attack. SimBA-DCT produces perturbations that are sparse in frequency space, and the resulting change in pixel space is spread out across all pixels.

\subsection{Google Cloud Vision attack}
\label{sec:gcv}

To demonstrate the efficacy of our attack against real world systems, we attack the Google Cloud Vision API, an online machine learning service that provides labels for arbitrary input images. For a given image, the API returns a list of top concepts contained in the image and their associated probabilities. Since the full list of probabilities associated with every label is unavailable, we define an untargeted attack that aims to remove the top 3 concepts in the original. We use the maximum of the original top 3 concepts' returned probabilities as the adversarial loss and use SimBA to minimize this loss. Figure \ref{fig:gcv_screenshot} shows a sample image before and after the attack. The original image (left) contains concepts related to camera instruments. SimBA successfully replaced the top concepts with weapon-related objects with imperceptible change to the original image. Additional samples are included in the supplementary material.

Since our attack can be executed efficiently, we evaluate its effectiveness over an aggregate of 50 random images. For the LFBA baseline, we define an attack as successful if the produced perturbation has an $L_2$-norm of at most the highest $L_2$-norm in a successful run of our attack. Figure \ref{fig:gcv_succ_plot} shows the average success rate of both attacks across number of queries. SimBA achieves a final success rate of $70\%$ after only 5000 API calls, while LFBA is able to succeed only $25\%$ of the time under the same query budge. To the best of our knowledge, this is the first adversarial attack result on Google Cloud Vision that has a high reported success rate within very limited number of queries.

\section{Related Work}
Many recent studies have shown that both white-box and black-box attacks can be applied to a diverse set of tasks. Computer vision models for image segmentation and object detection have also been shown to be vulnerable against adversarial perturbations \citep{cisse2017houdini, xie2017adversarial}. \citet{carlini2018audio} performed a systematic study of speech recognition attacks and showed that robust adversarial examples that alter the transcription model to output arbitrary target phrases can be constructed. Attacks on neural network policies \citep{huang2017adversarial, behzadan2017vulnerability} have also been shown to be permissible.

As these attacks become prevalent, many recent works have focused on designing defenses against adversarial examples. One common class of defenses applies an image transformation prior to classification, which aims to remove the adversarial perturbation without changing the image content \citep{xu2017feature, dziugaite2016study, guo2017countering}. Instead of requiring the model to correctly classify all adversarial images, another strategy is to detect the attack and output an adversarial class when certain statistics of the input appear abnormal \citep{li2017adversarial, metzen2017detecting, meng2017magnet, lu2017safetynet}. The training procedure can also be strengthened by including the adversarial loss as an implicit or explicit regularizer to promote robustness against adversarial perturbations \citep{tramer2017ensemble, madry2017towards, cisse2017parseval}. While these defenses have shown great success against a passive adversary, almost all of them can be easily defeated by modifying the attack strategy \citep{carlini2017bypass, athalye2018cvpr, athalye2018obfuscated}.

Relative to defenses against white-box attacks, few studies have focused on defending against adversaries that may only access the model via black-box queries. While transfer attacks can be effectively mitigated by methods such as ensemble adversarial training \citep{tramer2017ensemble} and image transformation \citep{guo2017countering}, it is unknown whether existing defense strategies can be applied to adaptive adversaries that may access the model via queries. \citet{guo2018low} have shown that the boundary attack is susceptible to image transformations that quantize the decision boundary, but employing the attack in low frequency space can successfully circumvent these transformation defenses.

\section{Discussion and Conclusion}
We proposed SimBA, a simple black-box adversarial attack that takes small steps iteratively guided by continuous-valued model output. The unprecedented query efficiency of our method establishes a strong baseline for future research on black-box adversarial examples. Given its real world applicability, we hope that more effort can be dedicated towards defending against malicious adversaries under this more realistic threat model.

While we intentionally avoid more sophisticated techniques to improve SimBA in favor of simplicity, we believe that additional modifications can still dramatically decrease the number of model queries. One possible extension could be to further investigate the selection of different sets of orthonormal bases, which could be crucial to the efficiency of our method by increasing the probability of finding a direction of large change. Another area for improvement is the adaptive selection of the step size $\epsilon$ to optimally consume the distance and query budgets.

Given that our method has very few requirements, it is conceptually suitable for applications to any task for which the target model returns a continuous score for the prediction. For instance, speech recognition systems are trained to maximize the probability of the correct transcription~\cite{amodei2016deepspeech2}, and policy networks~\cite{mnih2015human} are trained to maximize some reward function over the set of actions conditioned on the current environment. A simple iterative algorithm that modifies the input at random may prove to be effective in these scenarios. We leave these directions for future work.

\bibliography{main_icml2019}
\bibliographystyle{icml2019}

\newpage
\appendix
\onecolumn



\setcounter{table}{0}
\setcounter{figure}{0}
\renewcommand{\thesection}{S\arabic{section}}%
\renewcommand{\thetable}{S\arabic{table}}%
\renewcommand{\thefigure}{S\arabic{figure}}%

\title{Supplementary Material for Simple Black-box Adversarial Attacks}
\date{\vspace{-5ex}}

\maketitle

\section{Experiment on CIFAR-10}

In this section, we evaluate SimBA and SimBA-DCT on a ResNet-50 model trained on CIFAR-10. Both attacks remain very efficient on this new dataset without any hyperparameter tuning.

\begin{figure}[ht!]
\centering
\includegraphics[width=0.5\textwidth]{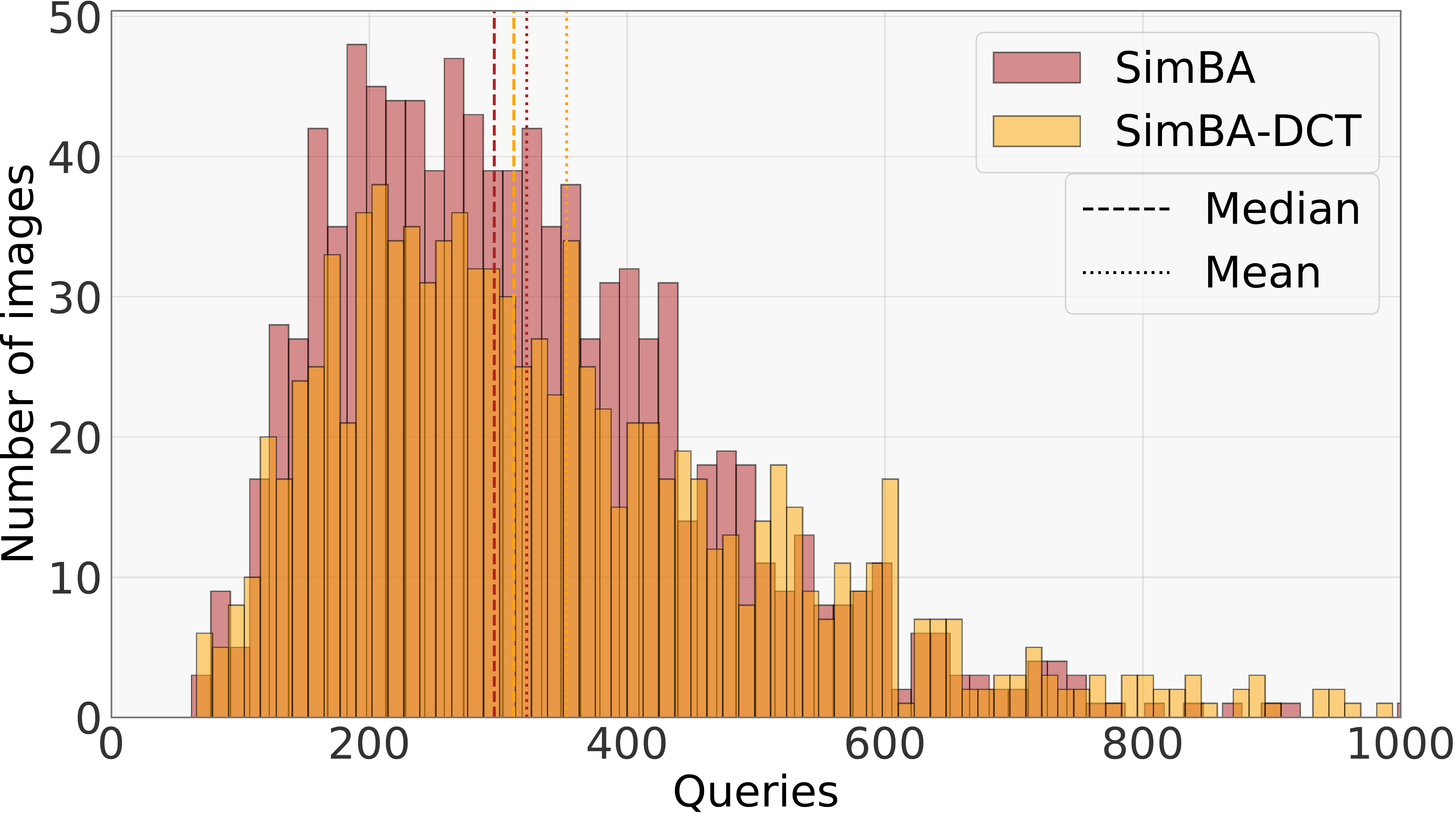}
\caption{Histogram of number of queries required until a successful targeted attack on CIFAR-10 (over 1000 target images).
\label{fig:queries_histogram_cifar}}
\end{figure}

\autoref{fig:queries_histogram_cifar} shows the distribution of queries required for a successful targeted attack to a random target label. In contrast to the experiment on ImageNet, the use of low frequency DCT basis is less effective due to the reduced image dimensionality. Both SimBA and SimBA-DCT perform similarly, with SimBA-DCT having a slightly heavier tail.

\autoref{table:method_comparison_cifar} shows aggregate statistics for the attack on CIFAR-10. Both methods achieve a success rate of $100\%$ when limited to a maximum of $10,000$ queries. The actual required queries is much fewer, with both methods averaging to approximately $300$ queries, matching the median. SimBA-DCT has a slightly worse performance compared to SimBA due its query distribution having a slightly heavier tail. Nevertheless, the average query count is in line with state-of-the-art attacks on CIFAR-10. For instance, AutoZOOM achieves a mean query count of $259$ with an average $L_2$-norm of 3.53.

\begin{table}[ht!]
\centering
\begin{tabular}{ccccc}
\hline
\textbf{Attack} & \textbf{Average queries} & \textbf{Median queries} & \textbf{Average $L_2$} & \textbf{Success rate} \\
\hline
\textbf{SimBA} & 322 & 297 & 2.04 & $100\%$ \\
\textbf{SimBA-DCT} & 353 & 312 & 2.21 & $100\%$ \\
\hline
\end{tabular}
\vspace{1ex}
\caption{Average query count for SimBA and SimBA-DCT on CIFAR-10.
\label{table:method_comparison_cifar}}
\end{table}

\section{Additional image samples for attack on Google Cloud Vision}

To demonstrate the generality of our evaluation of the Google Cloud Vision attack, we show 10 additional random images before and after perturbation by SimBA. In all cases, we successfully remove the top 3 original labels.

\begin{figure}[h!]
    \vspace{4ex}
    \centering
    \begin{subfigure}[]{0.9\textwidth}
    \includegraphics[width=\textwidth]{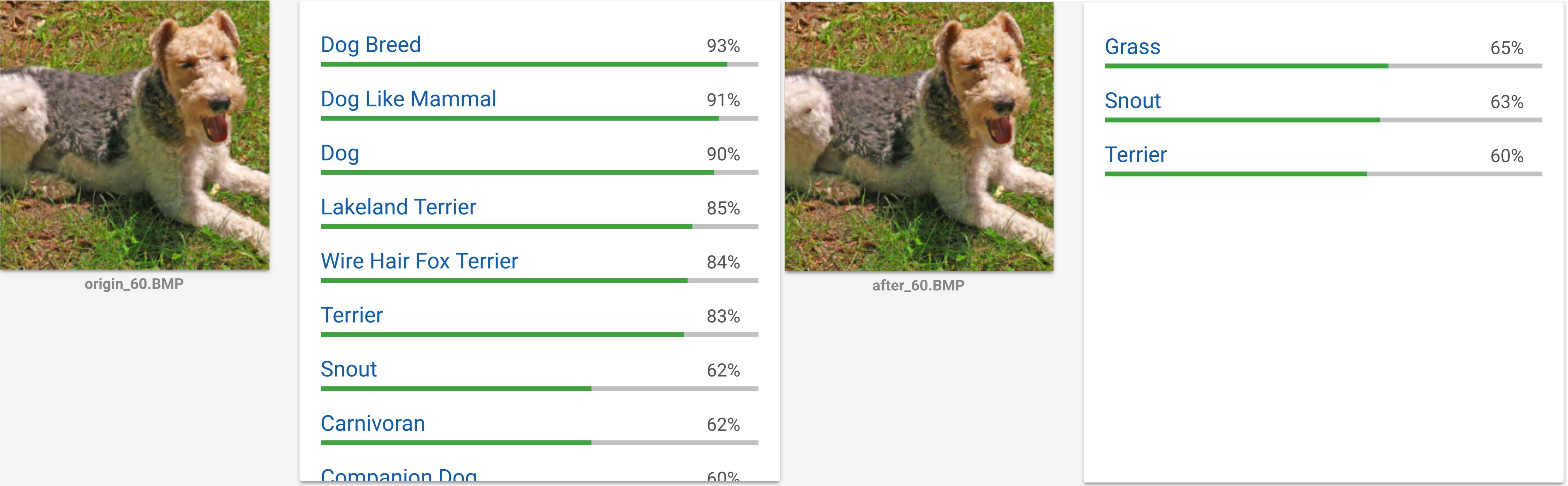}
    \end{subfigure}
    \begin{subfigure}[]{0.9\textwidth}
    \includegraphics[width=\textwidth]{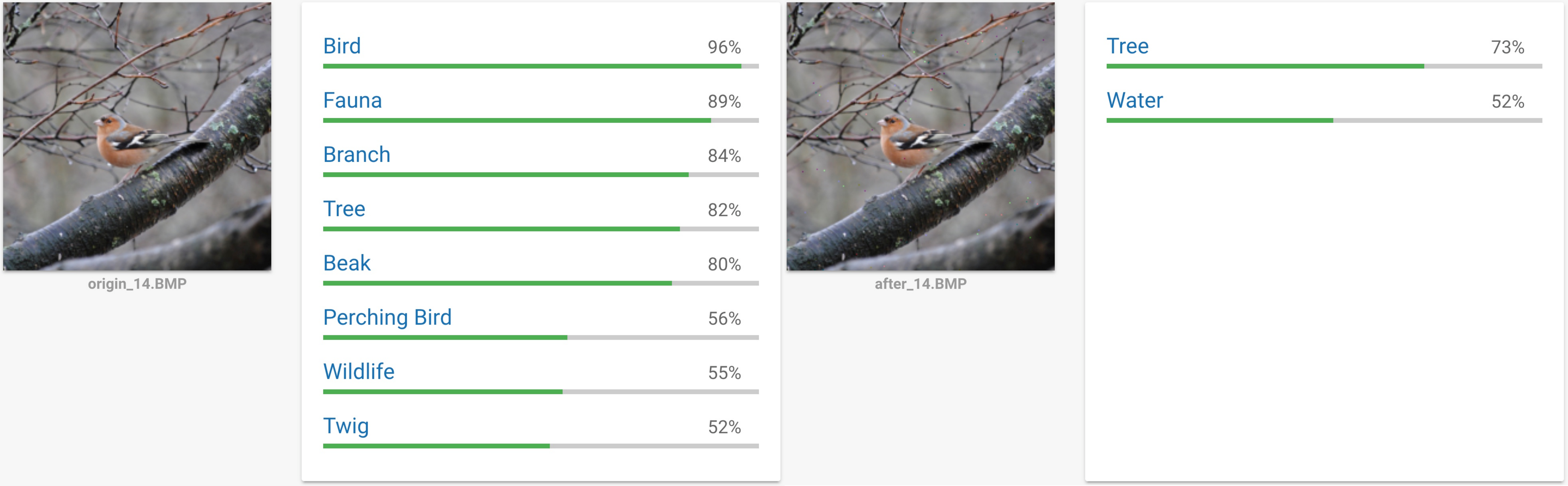}
    \end{subfigure}
    \begin{subfigure}[]{.9\textwidth}
    \includegraphics[width=\textwidth]{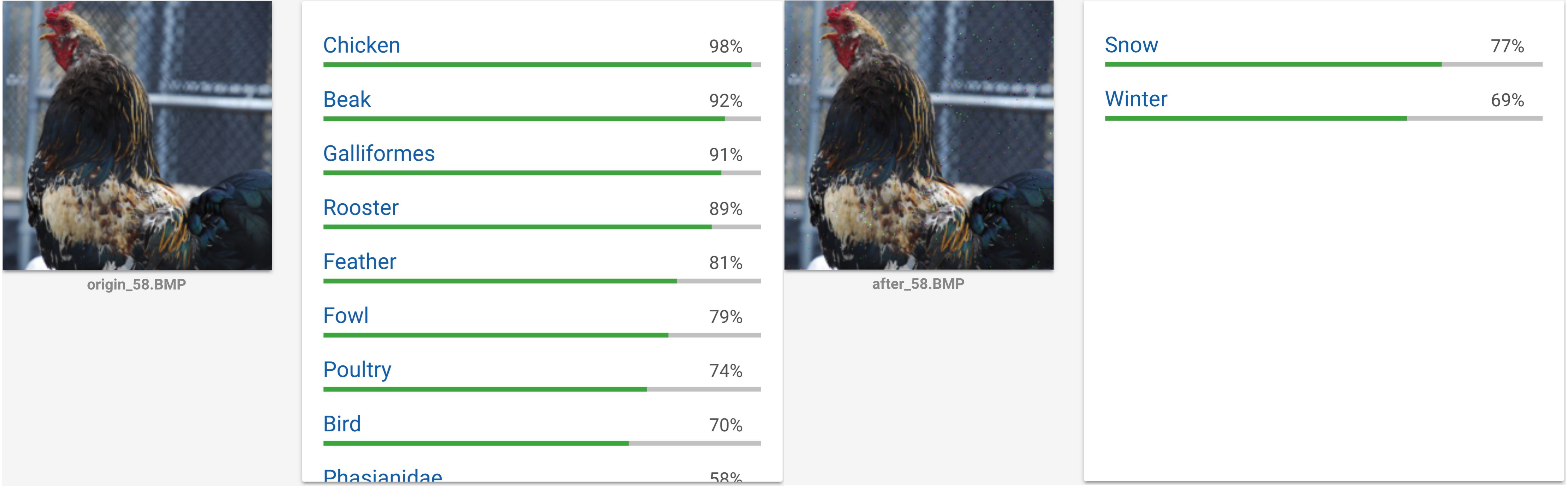}
    \end{subfigure}
    \begin{subfigure}[]{.9\textwidth}
    \includegraphics[width=\textwidth]{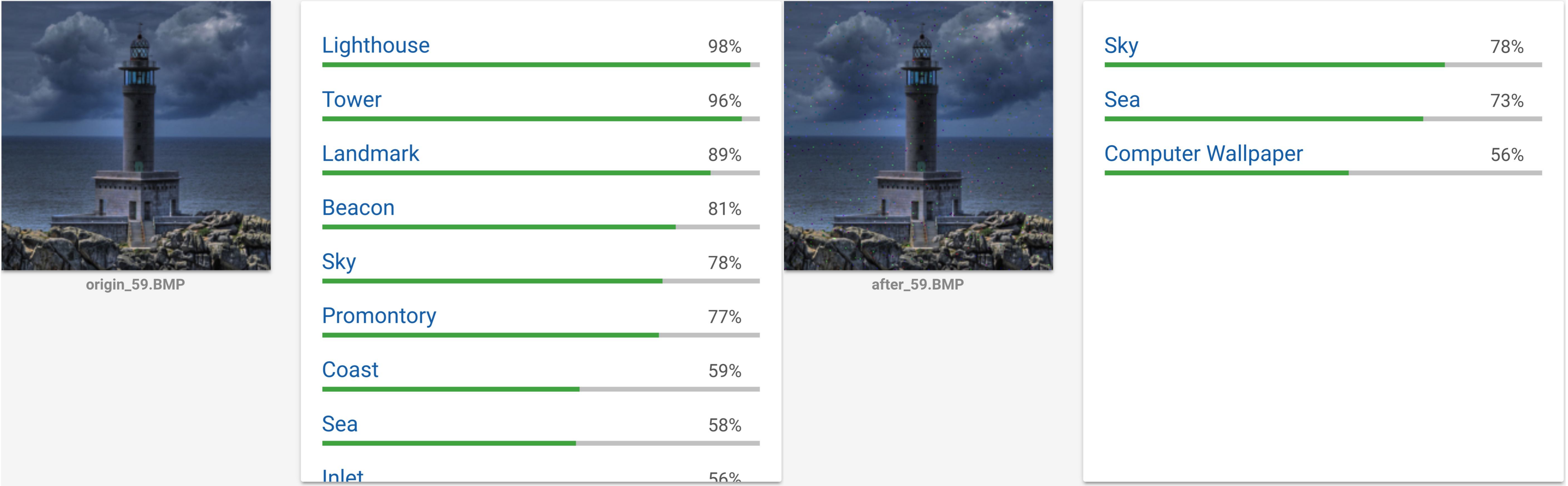}
    \end{subfigure}
\end{figure}

\begin{figure}[h!]
    \ContinuedFloat
    \centering
    \begin{subfigure}[]{.9\textwidth}
    \includegraphics[width=\textwidth]{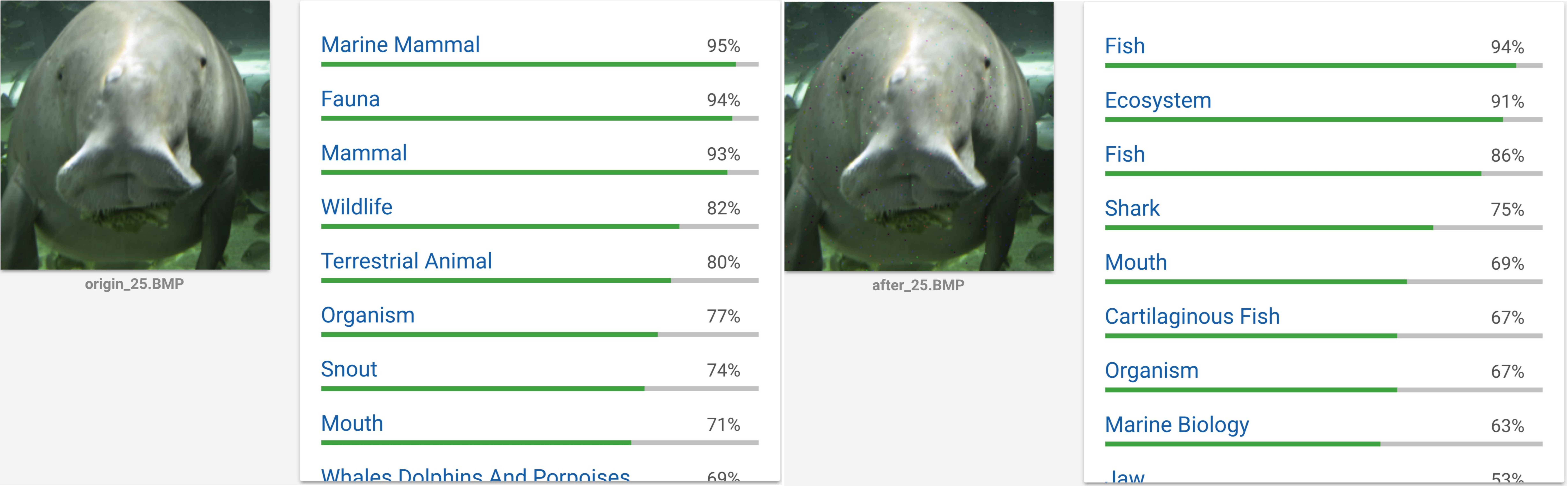}
    \end{subfigure}
    \begin{subfigure}[]{.9\textwidth}
    \includegraphics[width=\textwidth]{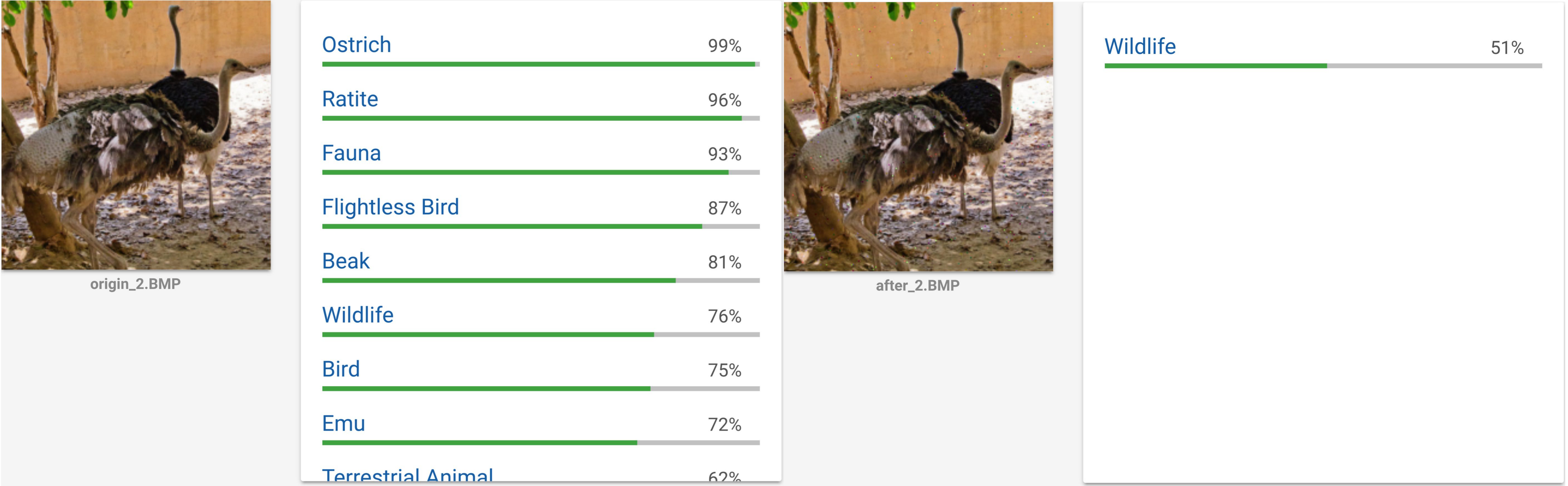}
    \end{subfigure}
    \begin{subfigure}[]{.9\textwidth}
    \includegraphics[width=\textwidth]{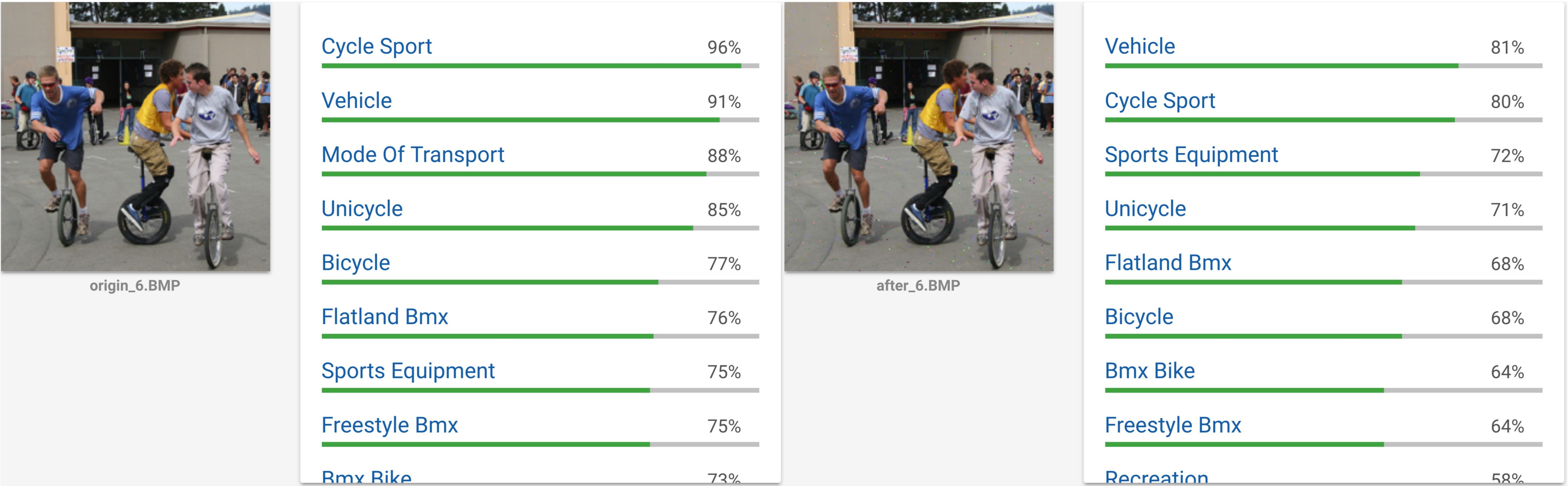}
    \end{subfigure}
    \begin{subfigure}[]{.9\textwidth}
    \includegraphics[width=\textwidth]{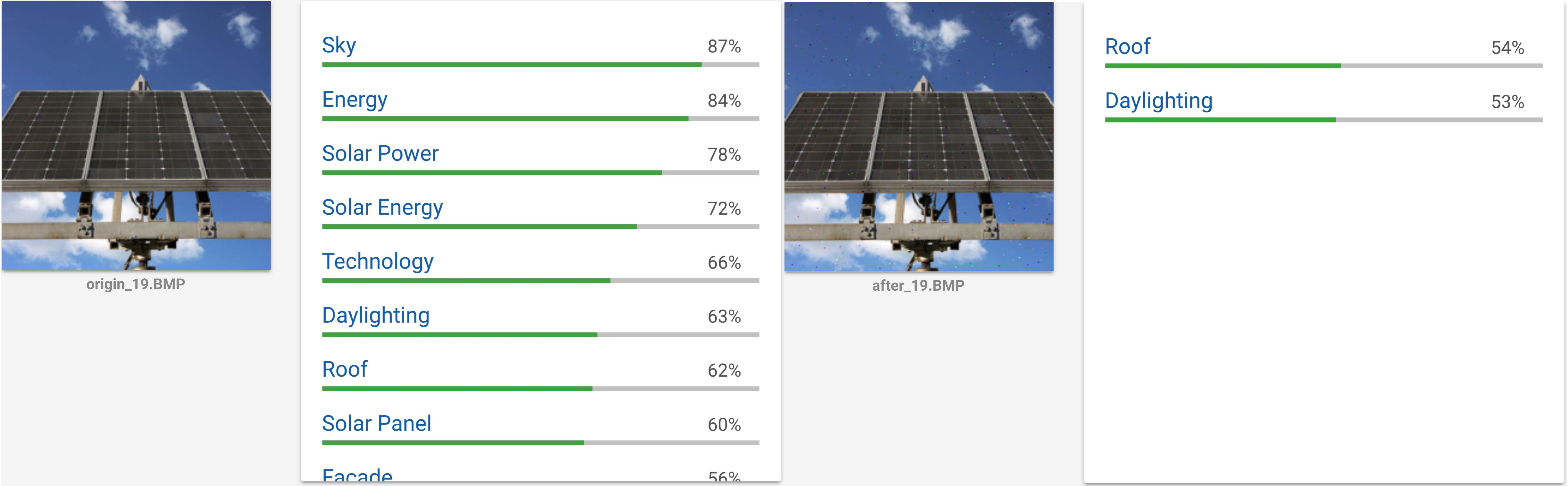}
    \end{subfigure}
\end{figure}

\begin{figure}[h!]
    \ContinuedFloat
    \centering
    \begin{subfigure}[]{.9\textwidth}
    \includegraphics[width=\textwidth]{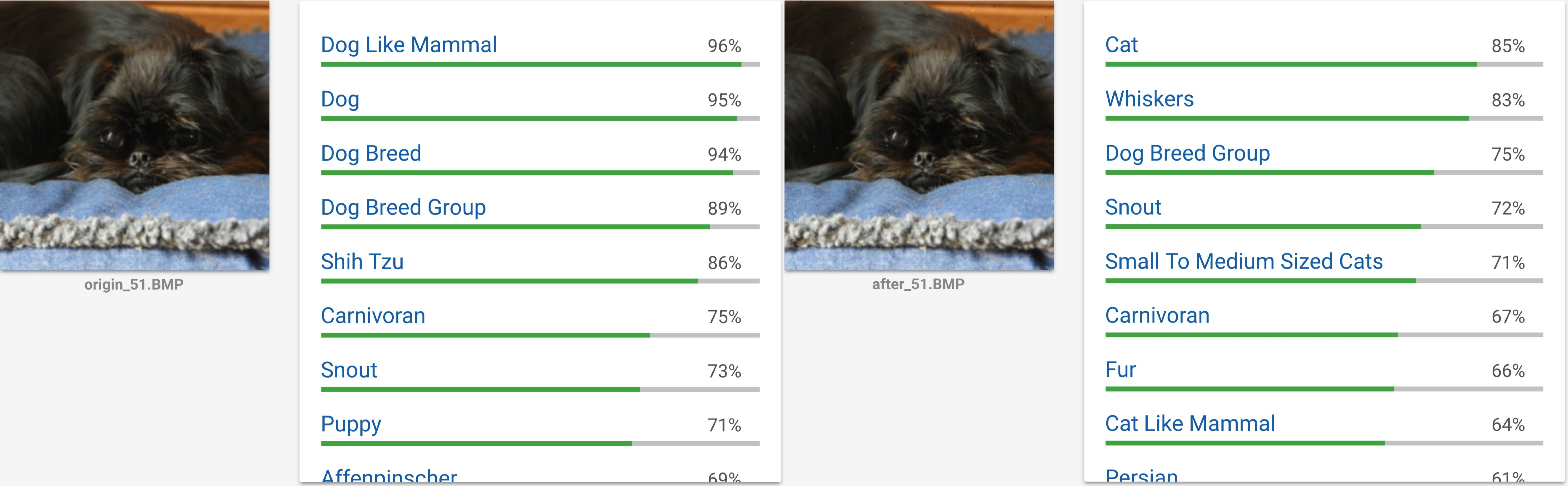}
    \end{subfigure}
    \begin{subfigure}[]{.9\textwidth}
    \includegraphics[width=\textwidth]{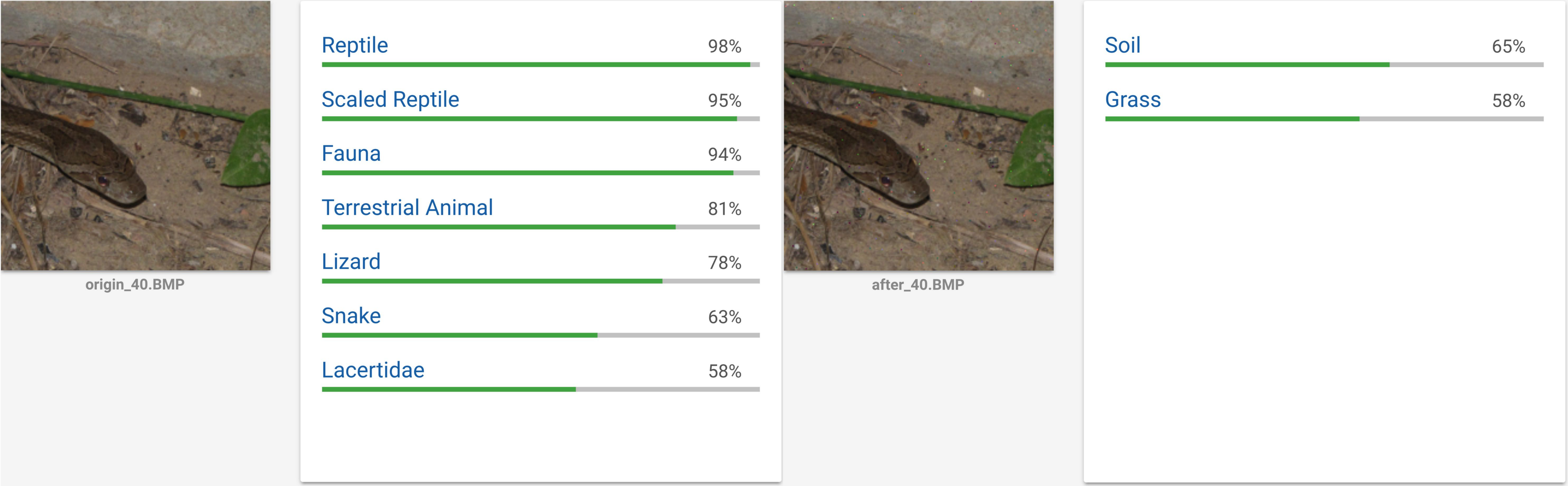}
    \end{subfigure}
    \caption{Additional adversarial images on Google Cloud Vision.}
\end{figure}


\end{document}